\documentclass[11pt]{article}
\usepackage{eacl2014}
\usepackage{times}
\usepackage{url}
\usepackage{latexsym}

\usepackage[lined,boxed,commentsnumbered]{algorithm2e}
\providecommand{\SetAlgoLined}{\SetLine} 
\usepackage{comment,multirow,caption,subcaption,graphicx}
\usepackage{float}
\floatstyle{plain}
\newfloat{algofloat}{thp}{lop}
\floatname{algofloat}{algofloat}

\title{Keyword and Keyphrase Extraction Using\\Centrality Measures on Collocation Networks}

\author{
Shibamouli Lahiri$^{1}$\hfil Sagnik Ray Choudhury$^{2}$\hfil Cornelia Caragea$^{1}$\\
$^{1}$Computer Science and Engineering\qquad $^{2}$Information Sciences and Technology\\
University of North Texas\hfil Pennsylvania State University\\
Denton, TX 76207, USA\hfil University Park, PA 16802, USA\\
{\tt shibamoulilahiri@my.unt.edu, szr163@ist.psu.edu, ccaragea@unt.edu}
}

\date{}

\begin{document}
\maketitle
\begin{abstract}
Keyword and keyphrase extraction is an important problem in natural language processing, with applications ranging from summarization to semantic search to document clustering. Graph-based approaches to keyword and keyphrase extraction avoid the problem of acquiring a large in-domain training corpus by applying variants of PageRank algorithm on a network of words. Although graph-based approaches are knowledge-lean and easily adoptable in online systems, it remains largely open whether they can benefit from centrality measures \emph{other than PageRank}. In this paper, we experiment with an array of centrality measures on word and noun phrase collocation networks, and analyze their performance on four benchmark datasets. Not only are there centrality measures that perform as well as or better than PageRank, but they are much simpler (e.g., \emph{degree}, \emph{strength}, and \emph{neighborhood size}). Furthermore, centrality-based methods give results that are competitive with and, in some cases, better than two strong unsupervised baselines.
\end{abstract}

\section{Introduction}
\label{sec:intro}

\begin{table}
\begin{center}
\footnotesize
\begin{tabular}{ccc}
\hline
\textbf{Degree} & \textbf{Strength} & \textbf{PageRank}\\
\hline
officials & white & officials\\
white & house & decision\\
decision & officials & white\\
official & official & official\\
house & senior & house\\
aid & decision & aid\\
senior & aid & coup\\
egypts & policy & senior\\
hagel & egypts & egypts\\
coup & hagel & hagel\\
\hline
\end{tabular}
\end{center}
\caption{\label{tab:degree_pagerank}Top 10 keywords selected by Degree, Strength (weighted degree) and PageRank on a sample newswire document on U.S.-Egypt diplomatic relations. Note that the three centrality measures return \emph{almost identical} top 10 words.}
\end{table}

Keyword and keyphrase extraction is the problem of automatically identifying {\em important} terms in text documents. These terms provide a high-level topic description of documents and are shown to be rich sources of information for applications such as: {\em online advertising}, i.e., displaying ads based on key terms extracted from webpages \cite{www200967}; {\em document summarization}, i.e., creating a summary with the most relevant points in a text \cite{Litvak:2008:GKE:1613172.1613178}; {\em clustering websites} \cite{1234007}; {\em tracking topics in newswire} \cite{Lee:2008:NKE:1443229.1444178}; {\em linking web documents to Wikipedia articles} \cite{Mihalcea:2007:WLD:1321440.1321475}; {\em detecting named entities} \cite{rennie2005using}; and {\em recommending academic papers} \cite{springer_link_cite}. Turney \shortcite{Turney:2003:CKE:1630659.1630724} observed that the problem had applications in ``summarizing, indexing, labeling, categorizing, clustering, highlighting, browsing, and searching''.

Supervised approaches to keyword and keyphrase extraction require large amounts of \emph{in-domain labeled data}, that are often expensive or impractical to acquire, especially for an online system where users can submit documents from any domain. Unsupervised approaches suffer from a similar problem - dependency on an external corpus or knowledge base (such as Wikipedia and WordNet) for computing \emph{phrase informativeness}~\cite{tomokiyo2003language,DBLP:conf/flairs/CsomaiM07,www200967}.

A class of unsupervised algorithms circumvents the above problems by employing variants of PageRank~\cite{Pageetal98} on co-occurrence networks of \emph{word types}~\cite{mihalcea-tarau:2004:EMNLP,Wan:2008:SDK:1620163.1620205}. These algorithms do not depend on an external knowledge source, and are therefore more suitable for deployment in online systems. An important point to note here is that PageRank is a \emph{centrality measure}, and returns the most central words in word co-occurrence networks.

Against this background, we ask the following question: \emph{Are there other centrality measures that perform as well as or better than PageRank in keyword/keyphrase extraction task?} Furthermore, are these centrality measures \emph{simpler} than PageRank? That this might indeed be the case, is evident from Table~\ref{tab:degree_pagerank}, where we show the top ten words selected by Degree, Strength (weighted degree), and PageRank on a sample word network of a newswire document. Note how similar the top-ranking words are for these three centrality measures. This calls into question whether we should use these other simpler centrality measures like Degree and Strength, and whether they are better than or competitive with PageRank on benchmark corpora. The research that we describe in this paper specifically addresses these questions.

We experiment with an array of centrality measures and their variants (cf. Table~\ref{tab:centrality_indices}) on different types of word and noun phrase collocation networks. Construction of these collocation networks has been described in Section~\ref{sec:collocation_networks}. These networks are then used to extract key terms from four benchmark datasets (cf. Section~\ref{sec:evaluation}). We treat word and noun phrase networks separately to have an idea about which centrality measures perform well on what types of networks. Note also that using noun phrase networks allows us to directly derive a ranking for phrases, as opposed to combining the ranks of constituent words from word networks.

{\bf Contributions and Organization.} Although centrality measures on word networks have been used before, there is no unifying study that looked systematically into different centrality measures on different types of collocation networks, in order to determine what works well and when. This is a gap we aim to bridge in our study. Our contributions are:
\begin{itemize}
\item Construction and experimentation with four different types of collocation networks on words and noun phrases (cf. Section~\ref{sec:collocation_networks})\footnote{Code and data available at \url{http://ec2-107-22-235-109.compute-1.amazonaws.com:8000}.}.
\item Ranking words and noun phrases based on eleven different centrality measures and their variations (cf. Section~\ref{sec:collocation_networks}).
\item Performance evaluation of different centrality measures and collocation networks on four benchmark datasets, and comparison of our method with \emph{tf-idf} -- a state-of-the-art unsupervised keyphrase extraction baseline (cf. Section~\ref{sec:evaluation}).
\item Identification of centrality measures and network types that perform well, and centrality measures that perform poorly (cf. Section~\ref{sec:evaluation}).
\item Design of an online graph-based keyword and keyphrase extraction system, based on the centrality measures used in this paper.\footnote{\url{http://ec2-107-22-235-109.compute-1.amazonaws.com:8000}.}
\end{itemize}

\section{Related Work}
\label{sec:related}

In the supervised paradigm, key term extraction is formulated as a \emph{classification} problem. Each term is encoded using different feature representations such as \emph{tf-idf}~\cite{Witten:1999:KPA:313238.313437,svmspringer_linkcite}, \emph{first occurrence}~\cite{Hulth:2003:IAK:1119355.1119383}, \emph{phrase length} and \emph{phrase distribution}~\cite{jiang2009ranking}, \emph{is-in-title}~\cite{Litvak:2008:GKE:1613172.1613178}, etc. Several machine learning algorithms have been used by different groups, e.g., na\"{i}ve Bayes by Frank et al.~\shortcite{Frank:1999:DKE:646307.687591}, and Witten et al.~\shortcite{Witten:1999:KPA:313238.313437}, decision trees by Ercan and Cicekli~\shortcite{Ercan:2007:ULC:1284916.1285164}, support vector machines by Zhang et al.~\shortcite{svmspringer_linkcite} and Jiang et al.~\shortcite{jiang2009ranking}, and conditional random fields by Zhang et al.~\shortcite{zhang2008automatic}.

In the unsupervised paradigm, the key term extraction problem is framed as a \emph{ranking} problem. Dominant ranking strategies include: \emph{tf}~\cite{Barker:2000:UNP:647461.726264}, \emph{tf-idf}~\cite{1234007}, \emph{term informativeness} and \emph{term phraseness}~\cite{tomokiyo2003language}, etc.

The performance of key term extraction systems is usually measured by Precision, Recall, and F-score. The state-of-the-art performance hovers around an F-score of 20\%-30\%~\cite{Kim:2010:STA:1859664.1859668}, thereby showing that key term extraction is a hard problem in general.


Graph-based methods to key term extraction are inherently unsupervised, where the philosophy is to build a network of words/phrases, and then rank the nodes using some kind of centrality measure. Researchers used variants of PageRank~\cite{mihalcea-tarau:2004:EMNLP,Wan:2008:SDK:1620163.1620205}, HITS~\cite{Litvak:2008:GKE:1613172.1613178}, and other measures like degree, betweenness and closeness~\cite{Xie:2005:CMT:1628960.1628980}. Use of noun phrase networks has been reported in~\cite{Xie:2005:CMT:1628960.1628980}.

While all these studies are extremely important, none of them systematically compared centrality measures in word and noun phrase collocation networks on benchmark corpora, in order to identify what measures work best and when. The studies by Xie~\shortcite{Xie:2005:CMT:1628960.1628980} and Palshikar~\shortcite{centrality_springer_link} are similar to our study, but there are significant differences. Xie~\shortcite{Xie:2005:CMT:1628960.1628980}, for example, was concerned with a similar but still different problem: predicting noun phrases that appear in paper abstracts, rather than entire documents. Palshikar~\shortcite{centrality_springer_link} did not test his approach on benchmark datasets, and did not report using different types of collocation networks and a suite of centrality measures.

Boudin~\shortcite{boudin:2013:IJCNLP} reported the first study on comparison of centrality measures for graph-based keyword extraction. Boudin's work is a pioneering study in its own right. While it may seem that our work is very similar to Boudin's, there are the following important differences:

\begin{itemize}
\item Boudin used only undirected graphs; we used both directed and undirected, and their simplified and non-simplified variants.
\item Boudin used syntactic filtering and stemming; we did not (to retain distinctions like ``learning'' and ``learnability'').
\item Boudin collapsed adjacent keywords into keyphrases. We generated keyphrases from noun phrase networks.
\item Boudin reported results on keyphrase extraction only; we report results on both keyword and keyphrase extraction.
\item Boudin used absolute cut-off for ranked keywords; we used percentage cut-off.
\item Boudin did not use as large a suite of centrality measures as we did. We tried 11 different centrality measures and their variations.
\item Boudin did not use as many benchmark datasets as we did, so our experimental results are more comprehensive. We used four different annotated datasets, and considered each annotation as a separate ground truth, resulting in 11 ground truths.
\end{itemize}

Finally, our experimental results are different than Boudin's, because our experimental set-up was different (no stemming, no syntactic filtering, no collapsing, etc).

\section{Collocation Networks and Centrality Measures}
\label{sec:collocation_networks}

Collocation networks\footnote{Also known as \emph{collocation graphs}~\cite{heyer2001learning,choudhury2009structure}.} are graphs where the nodes are unique words/phrases in a document, and edges link together two nodes if they occur within a certain window of each other~\cite{Ferret:2002:UCT:1072228.1072261}. Window sizes of two to ten words have been used by Mihalcea and Tarau~\shortcite{mihalcea-tarau:2004:EMNLP}.

\subsection{Word Collocation Networks}
\label{subsec:word_networks}

For word networks, we performed several text pre-processing steps. We lower-cased the input text, removed punctuation, numbers, stop words, and words with two characters or less. We did not perform stemming in order to retain subtle distinctions in text like ``learning'' and ``learnability''. Each document was converted to a collocation network as follows: nodes correspond to unique words in the document, and edges correspond to unique bigrams. More precisely, if word $w_1$ immediately preceded word $w_2$ in the pre-processed document, then an edge $w_1-w_2$ was added to the network. Bigram edges were specifically chosen after Mihalcea and Tarau \shortcite{mihalcea-tarau:2004:EMNLP}, who showed better performance for small window sizes in word collocation networks.

In our experiments, we used both weighted directed and weighted undirected graphs. The direction of an edge is from the first to the second word in a bigram, i.e., $w_1\rightarrow w_2$, and an edge weight is given by the bigram frequency in the document. Furthermore, we considered a simplified version of these graphs where all self-loops were removed, resulting in four types of collocation networks: directed, directed simplified, undirected, and undirected simplified (all weighted).

\subsection{Noun Phrase Collocation Networks}
\label{subsec:np_networks}

For noun phrase networks, we used an off-the-shelf sentence segmenter and noun phrase chunker available as part of the Apache OpenNLP toolkit\footnote{\tt{http://opennlp.apache.org/}.}. The chunker returns a list of unique noun phrases for each document. We filtered out noun phrases containing more than five words. A collocation network is constructed for each document as follows: nodes represent unique noun phrases, and edges link together noun phrases that occur within a specific window of each other. We chose the window size to be the median sentence length of a document. Similar to word networks, we constructed four types of collocation networks (as before). As a post-processing step, we removed all noun phrase vertices from the network whose words were of two characters or less. We also removed all single-word phrases that were stop words. We lower-cased the remaining noun phrases, and merged edges, when necessary. Note that the edges were all weighted with the co-occurrence frequency of $np_1$ and $np_2$. While merging edges, we simply added up the edge weights.

A na\"{i}ve algorithm for the construction of noun phrase collocation networks is to scan the whole document to determine how many times each pair of phrases in the phrase list occurs within a specified window size. However, this algorithm requires O($m^2$) passes (for $m$ phrases) over the document text, making it computationally very expensive. We instead made use of a sliding window algorithm that only did a \emph{single pass} through the entire document. It maintains a FIFO queue of \emph{longest prefix phrases}, and iteratively shrinks the prefix phrases by moving the window forward through the text.

\begin{table*}
\begin{center}
\scriptsize
\begin{tabular}{ll}
\hline
\textbf{Centrality Measure} & \textbf{Variations Explored}\\
\hline
Degree & in-degree, out-degree, degree\\
Strength (Weighted Degree) & in-strength, out-strength, strength\\
Neighborhood Size (order 1) & in-neighborhood size, out-neighborhood size, neighborhood size\\
Coreness & in-coreness, out-coreness, coreness\\
Clustering Coefficient & weighted, unweighted\\
Structural Diversity Index & N/A\\
PageRank & directed weighted, directed unweighted, undirected weighted, undirected unweighted\\
HITS & weighted and unweighted hub and authority scores\\
Betweenness & directed weighted, directed unweighted, undirected weighted, undirected unweighted\\
Closeness & weighted and unweighted in-closeness, out-closeness, closeness\\
Eigenvector Centrality & directed weighted, directed unweighted, undirected weighted, undirected unweighted\\
\hline
\end{tabular}
\end{center}
\caption{\label{tab:centrality_indices}Different centrality measures used in our study.}
\end{table*}

\subsection{Centrality Measures}
\label{subsec:centrality_measures}

\begin{figure*}
\centering
 \begin{subfigure}[b]{0.2396\textwidth}
 \centering
 \includegraphics[width=\textwidth]{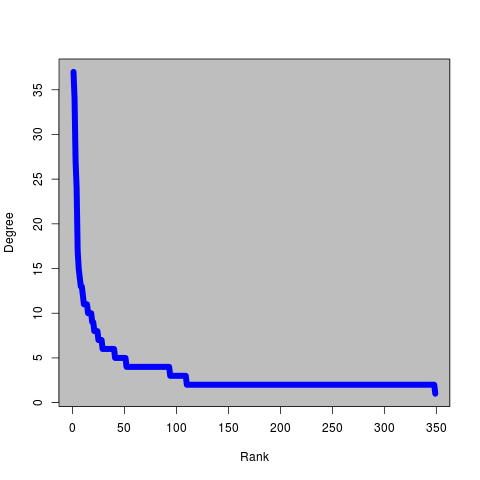}
 \caption{Degree}
 \label{fig:degree}
 \end{subfigure}
 \begin{subfigure}[b]{0.2396\textwidth}
 \centering
 \includegraphics[width=\textwidth]{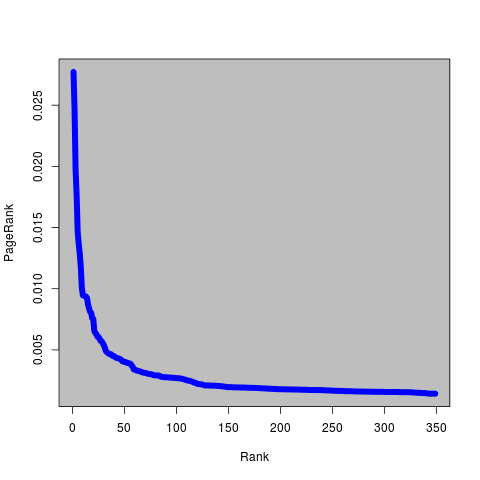}
 \caption{PageRank}
 \label{fig:pagerank}
 \end{subfigure}
 \begin{subfigure}[b]{0.2396\textwidth}
 \centering
 \includegraphics[width=\textwidth]{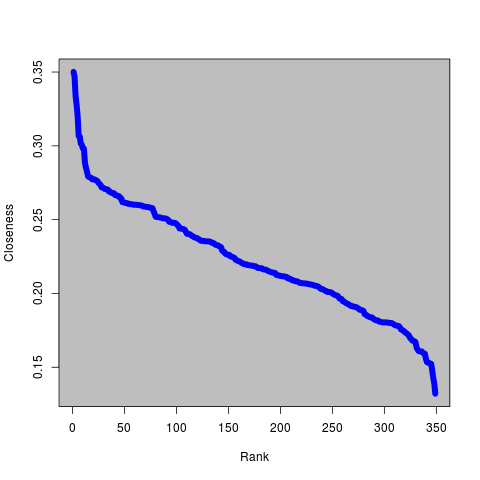}
 \caption{Closeness}
 \label{fig:closeness}
 \end{subfigure}
 \begin{subfigure}[b]{0.2396\textwidth}
 \centering
 \includegraphics[width=\textwidth]{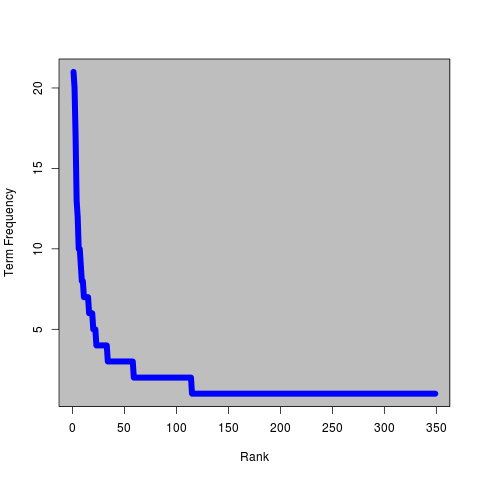}
 \caption{Term Frequency}
 \label{fig:tf}
 \end{subfigure}
\caption{Distributions of three centrality measures (degree, PageRank, and betweenness), and term frequency, on a word collocation network of a sample document from the NIPS collection of research papers.}
\label{fig:power_law}
\end{figure*}

To rank nodes in collocation networks, we used eleven different centrality measures and explored several of their variations, including weighted and unweighted versions of the graph whenever applicable\footnote{We used the \emph{igraph} package~\cite{igraph_cite} for our centrality computations.}. Moreover, each variation was applied to four different types of collocation networks - directed, directed simplified, undirected, and undirected simplified. Note that some of these variations yield identical rankings. For example, degree and neighborhood size (order 1) are identical on a simplified network without self-loops. Here we treat identical rankings separately for the sake of conceptual clarity. Also note that Structural Diversity Index~\cite{citeulike:7205422} is not a centrality measure, but behaves similarly if we sort vertices in the increasing order of diversity. Table~\ref{tab:centrality_indices} lists all centrality measures and their variations used in our study. Definitions of the centrality measures follow:\\
\textbf{1. Degree:} number of edges incident to a node.\\
\textbf{2. Strength:} sum of the weights of the edges incident to a node.\\
\textbf{3. Neighborhood size - order 1:} number of immediate neighbors to a node.\\
\textbf{4. Coreness:} outermost core number of a node in the $k$-core decomposition of a graph~\cite{ICT4DBibliography2427,DBLP:journals/corr/cs-DS-0310049}.\\
\textbf{5. Clustering Coefficient:} density of edges among the immediate neighbors of a node~\cite{watts1998cds}.\\
\textbf{6. Structural Diversity Index:} normalized entropy of the weights of the edges incident to a node~\cite{citeulike:7205422}.\\
\textbf{7. PageRank:} importance of a node based on how many important nodes it is connected to~\cite{Pageetal98}.\\
\textbf{8. HITS:} importance of a node as a \emph{hub} (pointing to many others) and as an \emph{authority} (pointed to by many others)~\cite{Kleinberg:1999:ASH:324133.324140}.\\
\textbf{9. Betweenness:} fraction of shortest paths that pass through a node, summed over all node pairs~\cite{citeulike:8254181,doi:10.1080/0022250X.2001.9990249}.\\
\textbf{10. Closeness:} reciprocal of the sum of distances of all nodes to some node~\cite{Bavelas:1950}.\\
\textbf{11. Eigenvector Centrality:} element of the first eigenvector of a graph adjacency matrix corresponding to a node~\cite{Bonacich87}.

Figure \ref{fig:power_law} shows the distributions of three centrality measures - degree, PageRank, and closeness, as well as the distribution of term frequency on a word collocation network of a sample document from the NIPS collection of research articles\footnote{\url{http://www.cs.nyu.edu/~roweis/data.html}}. As can be seen from the figure, the network exhibits power-law distributions for centrality measures (Figures \ref{fig:degree}, \ref{fig:pagerank}, and \ref{fig:closeness}), relating them to the Zipfian curve of term frequency in Figure~\ref{fig:tf}~\cite{Zip35}. This is not surprising since in a small-world network, there are a few nodes that are connected to many others, and many nodes that are connected to only a few close neighbors~\cite{citeulike:44_cn}. Moreover, as Ruhnau \shortcite{Ruhnau2000357} pointed out, centrality measures form equivalence classes such as \emph{node centrality} and \emph{point centrality}, indicating that their empirical distributions should be very similar, if not the same.

Since many centrality measures exhibit a power-law distribution, researchers proposed choosing the most important terms that appear above the ``knee'' of those distributions~\cite{www200967}. Grineva et al.~\shortcite{www200967} mentioned that this \emph{decline}-based thresholding leads to higher precision in keyphrase extraction, especially in the presence of noisy and multi-theme documents. However, this decline-based thresholding is not always feasible, because it is not clear where the ``knee'' of the curve occurs. For example, Figure~\ref{fig:degree} has a clear knee, whereas Figure~\ref{fig:closeness} does not.

We therefore chose to use the more traditional percentage-based thresholding, where we rank words and noun phrases based on their centrality, and then return the top $k\%$ of words/phrases as our list of keywords/phrases\footnote{This strategy has the added advantage that users can choose to return as many key terms as they want. It was also adopted in~\cite{mihalcea-tarau:2004:EMNLP}.}.

\section{Evaluation}
\label{sec:evaluation}

\begin{table*}
\begin{center}
\scriptsize
\begin{tabular}{lccc|ccc}
\hline
\multirow{4}{*}{\textbf{Benchmark}} & \multirow{2}{*}{\textbf{Best Centrality-based}} & \multirow{2}{*}{\textbf{Best tf-based}} & \multirow{2}{*}{\textbf{Best tf-idf-based}} & \textbf{Best Centrality-based} & \textbf{tf-based} & \textbf{tf-idf-based}\\
& \multirow{2}{*}{\textbf{F-score (\%)}} & \multirow{2}{*}{\textbf{F-score (\%)}} & \multirow{2}{*}{\textbf{F-score (\%)}} & \textbf{F-score (\%)} & \textbf{F-score (\%)} & \textbf{F-score (\%)}\\
&&&& \textbf{on Average} & \textbf{on Average} & \textbf{on Average}\\
& \multicolumn{3}{c|}{(top $k$\% to achieve these F-scores are in parentheses)} & \multicolumn{3}{c}{(standard deviations are in parentheses)}\\
\hline
ICSI - 1 & 12.45 (5) & 12.04 (5) & \textbf{17.8 (5)} & 4.62 (2.91) & 4.58 (2.82) & \textbf{5.13 (3.97)}\\
ICSI - 2 & 11.0 (5) & 11.0 (5) & \textbf{16.84 (5)} & 4.95 (2.41) & 4.89 (2.39) & \textbf{5.73 (3.55)}\\
ICSI - 3 & 9.22 (5) & 8.79 (5) & \textbf{12.54 (5)} & 3.31 (1.98) & 3.27 (1.91) & \textbf{3.79 (2.68)}\\
ICSI - combined & 17.44 (5) & 17.03 (5) & \textbf{26.16 (5)} & 8.67 (3.97) & 8.63 (3.85) & \textbf{9.98 (5.69)}\\
NUS - combined & 6.85 (5) & 6.75 (5) & \textbf{7.31 (5)} & 1.69 (1.52) & 1.68 (1.49) & \textbf{1.71 (1.61)}\\
INSPEC - controlled & 4.16 (10) & 4.13 (10) & \textbf{4.2 (10)} & 2.45 (0.78) & 2.46 (0.78) & \textbf{2.55 (0.74)}\\
INSPEC - uncontrolled & 7.2 (5) & 6.63 (10) & \textbf{8.67 (5)} & 5.29 (0.98) & 4.98 (0.82) & \textbf{6.18 (1.43)}\\
INSPEC - combined & 8.97 (15) & 8.72 (15) & \textbf{10.46 (10)} & 6.73 (1.18) & 6.49 (1.09) & \textbf{7.7 (1.6)}\\
SemEval - author & \textbf{2.68 (5)} & 2.62 (5) & \textbf{2.68 (5)} & 0.57 (0.58) & 0.57 (0.57) & \textbf{0.58 (0.59)}\\
SemEval - reader & \textbf{4.64 (5)} & 4.51 (5) & 4.45 (5) & \textbf{1.1 (1.0)} & \textbf{1.1 (0.99)} & \textbf{1.1 (0.98)}\\
SemEval - combined & \textbf{6.32 (5)} & 6.16 (5) & 6.11 (5) & 1.48 (1.36) & 1.48 (1.35) & \textbf{1.49 (1.35)}\\
\hline
\end{tabular}
\end{center}
\caption{\label{tab:keyword_best_tf_tfidf} Comparison of the best F-scores between centrality-based rankings and {\em tf} and {\em tf-idf} based rankings for {\bf keyword extraction}, on all four datasets, for each of their associated gold standard annotations. We also report best F-scores obtained \emph{on average}, along with standard deviations, where the average was taken across top $k$\% thresholds, with $k$ from 5 to 100, in steps of 5. Best performance is highlighted in each row.
}
\end{table*}

\begin{figure*}
\centering
 \begin{subfigure}[b]{0.24\textwidth}
 \centering
 \includegraphics[width=\textwidth]{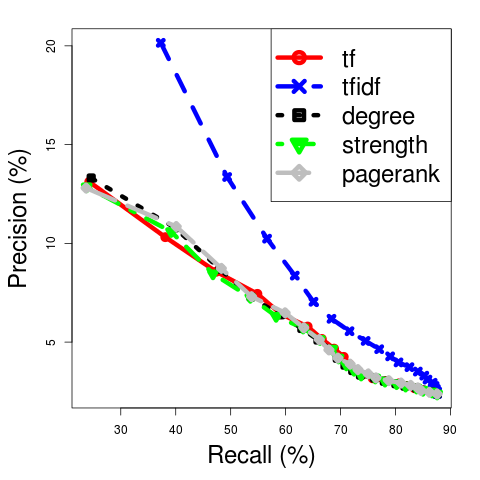}
 \caption{ICSI Keywords}
 \label{fig:ICSI_keywords}
 \end{subfigure}
 \begin{subfigure}[b]{0.24\textwidth}
 \centering
 \includegraphics[width=\textwidth]{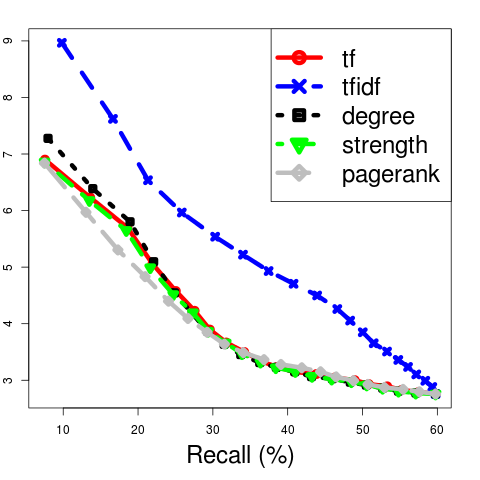}
 \caption{INSPEC Keywords}
 \label{fig:INSPEC_keywords}
 \end{subfigure}
 \begin{subfigure}[b]{0.24\textwidth}
 \centering
 \includegraphics[width=\textwidth]{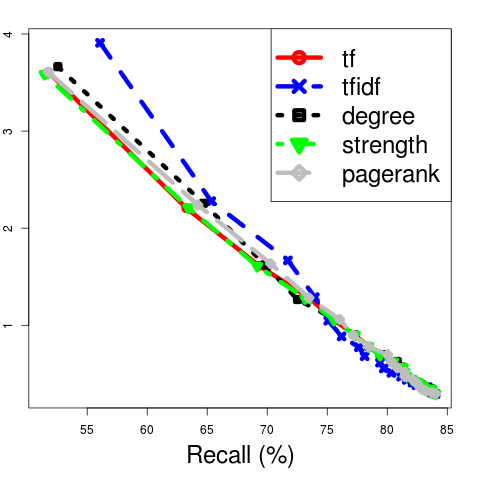}
 \caption{NUS Keywords}
 \label{fig:NUS_keywords}
 \end{subfigure}
 \begin{subfigure}[b]{0.24\textwidth}
 \centering
 \includegraphics[width=\textwidth]{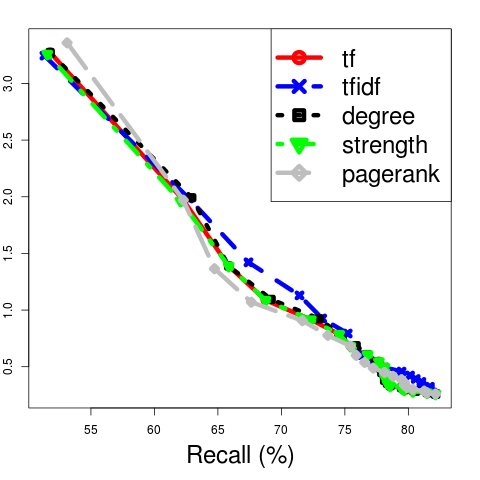}
 \caption{SemEval Keywords}
 \label{fig:SemEval_keywords}
 \end{subfigure}
\caption{Precision-Recall curves for keyword extraction, using the \emph{combined} gold standard annotations.}
\label{fig:keyword_PR}
\end{figure*}

\begin{table*}
\begin{center}
\scriptsize
\begin{tabular}{lccc|ccc}
\hline
\multirow{4}{*}{\textbf{Benchmark}} & \multirow{2}{*}{\textbf{Best Centrality-based}} & \multirow{2}{*}{\textbf{Best tf-based}} & \multirow{2}{*}{\textbf{Best tf-idf-based}} & \textbf{Best Centrality-based} & \textbf{tf-based} & \textbf{tf-idf-based}\\
& \multirow{2}{*}{\textbf{F-score (\%)}} & \multirow{2}{*}{\textbf{F-score (\%)}} & \multirow{2}{*}{\textbf{F-score (\%)}} & \textbf{F-score (\%)} & \textbf{F-score (\%)} & \textbf{F-score (\%)}\\
&&&& \textbf{on Average} & \textbf{on Average} & \textbf{on Average}\\
& \multicolumn{3}{c|}{(top $k$\% to achieve these F-scores are in parentheses)} & \multicolumn{3}{c}{(standard deviations are in parentheses)}\\
\hline
ICSI - 1 & \textbf{9.7 (5)} & 7.57 (5) & 8.17 (5) & \textbf{3.65 (2.24)} & 3.28 (1.73) & 3.14 (1.82)\\
ICSI - 2 & 7.85 (5) & 6.42 (5) & \textbf{8.08 (5)} & \textbf{3.34 (1.87)} & 3.09 (1.45) & 3.1 (1.75)\\
ICSI - 3 & \textbf{6.83 (5)} & 4.6 (5) & 4.99 (5) & \textbf{2.3 (1.45)} & 1.95 (0.97) & 1.83 (1.04)\\
ICSI - combined & \textbf{11.1 (10)} & 8.93 (10) & 9.54 (5) & \textbf{5.57 (2.51)} & 4.95 (1.83) & 4.83 (2.02)\\
NUS - combined & 6.42 (5) & 6.42 (5) & \textbf{7.65 (5)} & \textbf{2.78 (1.51)} & 2.68 (1.42) & 2.74 (1.68)\\
INSPEC - controlled & \textbf{1.71 (40)} & 1.47 (55) & 1.49 (90) & \textbf{1.55 (0.15)} & 1.35 (0.16) & 1.26 (0.16)\\
INSPEC - uncontrolled & 12.06 (100) & 12.06 (100) & \textbf{12.79 (80)} & 9.96 (2.31) & 9.15 (2.53) & \textbf{10.24 (2.74)}\\
INSPEC - combined & 10.88 (100) & 10.88 (100) & \textbf{11.28 (85)} & 8.58 (2.24) & 7.85 (2.45) & \textbf{8.72 (2.63)}\\
SemEval - author & \textbf{3.24 (5)} & 2.57 (5) & 2.97 (5) & \textbf{0.94 (0.71)} & 0.9 (0.58) & 0.91 (0.66)\\
SemEval - reader & \textbf{4.21 (5)} & 3.55 (5) & 4.12 (5) & \textbf{1.89 (0.9)} & 1.84 (0.74) & 1.88 (0.88)\\
SemEval - combined & \textbf{5.51 (5)} & 4.35 (5) & 5.1 (5) & \textbf{2.38 (1.19)} & 2.28 (0.93) & 2.32 (1.11)\\
\hline
\end{tabular}
\end{center}
\caption{\label{tab:keyphrase_best_tf_tfidf} Comparison of the best F-scores between centrality-based rankings and {\em tf} and {\em tf-idf} based rankings for {\bf keyphrase extraction}, on all four datasets, for each of their associated gold standard annotations. We also report best F-scores obtained \emph{on average}, along with standard deviations, where the average was taken across top $k$\% thresholds, with $k$ from 5 to 100, in steps of 5. Best performance is highlighted in each row.
}
\end{table*}

\begin{figure*}
\centering
 \begin{subfigure}[b]{0.24\textwidth}
 \centering
 \includegraphics[width=\textwidth]{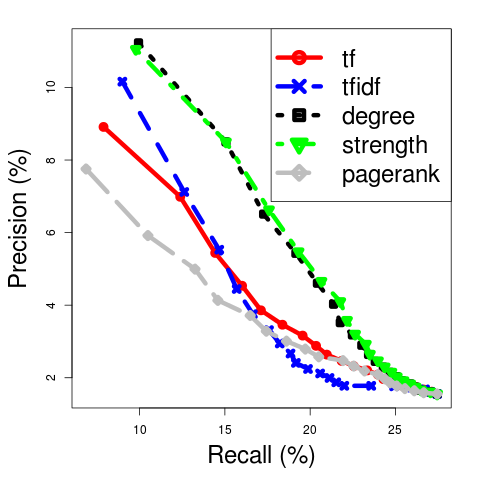}
 \caption{ICSI Keyphrases}
 \label{fig:ICSI_keyphrases}
 \end{subfigure}
 \begin{subfigure}[b]{0.24\textwidth}
 \centering
 \includegraphics[width=\textwidth]{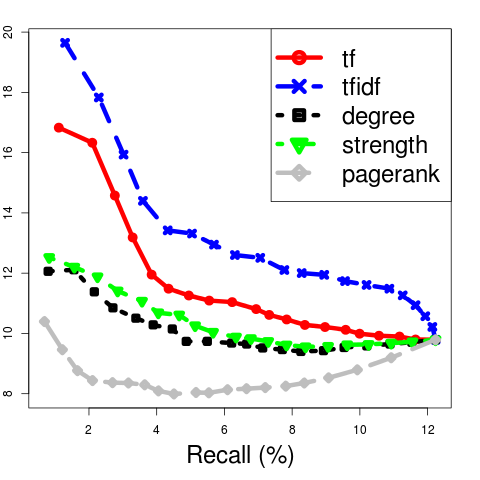}
 \caption{INSPEC Keyphrase}
 \label{fig:INSPEC_keyphrases}
 \end{subfigure}
 \begin{subfigure}[b]{0.24\textwidth}
 \centering
 \includegraphics[width=\textwidth]{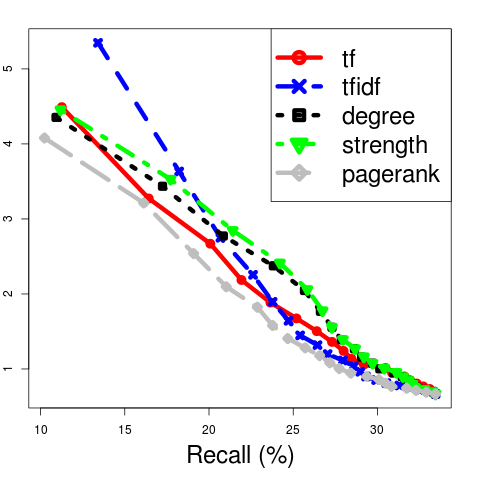}
 \caption{NUS Keyphrases}
 \label{fig:NUS_keyphrases}
 \end{subfigure}
 \begin{subfigure}[b]{0.24\textwidth}
 \centering
 \includegraphics[width=\textwidth]{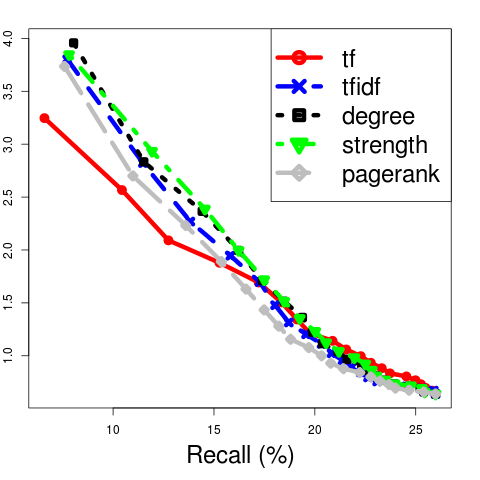}
 \caption{SemEval Keyphrase}
 \label{fig:SemEval_keyphrases}
 \end{subfigure}
 \caption{Precision-Recall curves for keyphrase extraction, using the \emph{combined} gold standard annotations.}
\label{fig:keyphrase_PR}
\end{figure*}

We used four different benchmark datasets to evaluate our centrality-based ranking strategies in collocation networks. The first three datasets are described in~\cite{hasan2010conundrums}, and the fourth dataset is described in~\cite{Kim:2010:STA:1859664.1859668}. The datasets are as follows:\\
\textbf{ICSI} consists of a collection of meeting transcripts divided into 201 segments~\cite{janin2003icsi,Liu:2009:UAA:1620754.1620845}. We removed all segments marked as ``chitchat'' and ``digit'' (after~\cite{Liu:2009:UAA:1620754.1620845} and~\cite{hasan2010conundrums}), ending up with 160 segments. Three sets of independent gold standard annotations are available for each segment. In addition, we created a fourth set by taking the union of all three.\\
\textbf{NUS} consists of a set of 211 academic papers~\cite{Nguyen:2007:KES:1780653.1780707}. Following the strategy specified in~\cite{hasan2010conundrums}, we generated a single set of gold standard annotations for this dataset. For each paper, we removed the title, authors, affiliations and references as a pre-processing step.\\
\textbf{INSPEC} consists of 2000 abstracts from journal papers, including paper titles~\cite{Hulth:2003:IAK:1119355.1119383,mihalcea-tarau:2004:EMNLP}. Two sets of gold standard annotations, one with a \emph{controlled vocabulary}, and the other with an \emph{uncontrolled vocabulary}, are available for each abstract. We created a third gold standard by merging them.\\
\textbf{SemEval} consists of 144 academic papers that constituted the training dataset in SemEval 2010 Keyphrase Extraction Task~\cite{Kim:2010:STA:1859664.1859668}. We excluded the test set because the unstemmed keyphrases on the test set have not been released\footnote{Note that since we constructed word and noun phrase networks on unstemmed lexical units, our systems require that we compare the returned keywords/keyphrases against unstemmed gold standard annotations. This is a general strategy we followed throughout all our experiments.}. We used three sets of gold standard annotations available for this set, the \emph{author-assigned keyphrases}, the \emph{reader-assigned keyphrases}, and the \emph{combined keyphrases}.

To evaluate our centrality-based rankings, we resorted to three popular metrics used in keyphrase extraction - precision, recall, and F-score. Note that for word networks, we compare our results against the unigram-subset of the gold standard annotations as many of the keyphrases in these datasets are actually single words~\cite{hasan2010conundrums}. For noun phrase networks, we used the complete set of gold standard annotations.

\subsection{Keyword Extraction}
\label{sec:keyword_extraction}

Table \ref{tab:keyword_best_tf_tfidf} shows the comparison of F@$k$\% between centrality-based rankings and {\em tf} and {\em tf-idf} based rankings for keyword extraction, on the four datasets used in this study, for each of their associated gold standard annotations. The best F-scores and the corresponding top $k$\% are given for each annotation separately. We tested the percentage-based thresholding strategy on top 5\% to top 100\% (at an interval of 5) of the total number of words. Table~\ref{tab:keyword_best_tf_tfidf} shows, along with the best F-score values, the corresponding thresholding percentage in parentheses. In addition, in Table \ref{tab:keyword_best_tf_tfidf}, we also display the best F-score values obtained after we take the average of F@$k$\% for all $k$ from 5 to 100, at an interval of 5.

Hasan and Ng \shortcite{hasan2010conundrums} showed that {\em tf-idf} often outperformed many existing approaches. Thus, we compared the centrality-based ranking with the following two baselines: a weak term frequency baseline, and the much stronger {\em tf-idf} baseline. The {\em idf} component was computed from the whole corpus. For example, while extracting keywords from an ICSI document, we extracted the term frequencies from that document, and extracted the {\em idfs} from the whole ICSI corpus.

As can be seen from Table~\ref{tab:keyword_best_tf_tfidf}, the centrality-based rankings outperform the weak term frequency baseline most of the times (for both the best F-score and the F-score on average), but it only outperforms the stronger {\em tf-idf} baseline (for best F-score) on the SemEval dataset. On average, centrality rankings perform similarly with or slightly worse than the {\em tf-idf} baseline. This shows that {\em tf-idf} is indeed a very simple and effective ranking strategy~\cite{hasan2010conundrums}. Note, however, that computing the {\em idf} component requires the use of an \emph{external corpus}, and in a real-life scenario, a user in need of extracting keywords from a particular document, may not be able to specify such a corpus off-hand. This is where the centrality-based rankings have greater applicability. Moreover, centrality measures are very fast and easy to compute on single-document graphs, as compared to other more complex unsupervised ranking strategies, e.g., the mixture model likelihood introduced by~\cite{rennie2005using}.

Another important observation is that the best performers in keyword extraction from collocation networks are variants of {\em degree}, {\em strength}, {\em PageRank}, and {\em neighborhood size (order 1)}, whereas the worst performers are the {\em Structural Diversity Index}~\cite{citeulike:7205422}, and variants of {\em clustering coefficient} (table included in the supplementary material).

Figures~\ref{fig:ICSI_keywords},~\ref{fig:INSPEC_keywords},~\ref{fig:NUS_keywords} and~\ref{fig:SemEval_keywords} display the Precision-Recall curves for the keyword extraction task. Here we used directed graphs and the \emph{combined} gold standard annotations. We ranked vertices using {\em tf}, {\em tf-idf}, {\em degree}, {\em strength}, and {\em PageRank} on weighted graphs, and again, used threshold values of 5\% to 100\% in steps of 5. 
As can be seen from the figure, {\em tf-idf} achieves a much higher precision for the same recall as compared with the other four rankings for ICSI and INSPEC, and have similar precision for the same recall among all rankings for NUS and SemEval. PageRank performs very similarly, or slightly worse, when compared with {\em degree} and {\em strength} based rankings on all datasets.

\subsection{Keyphrase Extraction}
\label{sec:keyphrase_extraction}

Table~\ref{tab:keyphrase_best_tf_tfidf} is structurally similar to Table \ref{tab:keyword_best_tf_tfidf}, and it shows the results for {\em keyphrase extraction} on different gold standard annotations. It is interesting to note that in seven out of eleven gold standard annotations, the best centrality-based F-scores outperformed the strong {\em tf-idf} baseline, and in nine out of eleven, they beat the {\em tf-idf} baseline \emph{on average}. This shows the potential of centrality-based rankings in keyphrase extraction. As already mentioned, {\em tf-idf} requires an external corpus to compute the {\em idf} component. Here, we outperform {\em tf-idf} \emph{without} using any external corpus. Furthermore, centrality-based rankings in Table~\ref{tab:keyphrase_best_tf_tfidf} yield better F-score values on the NUS and ICSI datasets than the ones obtained with TextRank, SingleRank, ExpandRank and KeyCluster algorithms in~\cite{hasan2010conundrums} (cf. Table 2 in~\cite{hasan2010conundrums}). Centrality measures also beat the weak term frequency baseline on most of the gold standards. This time, the best performers were variants of {\em degree}, {\em neighborhood size (order 1)}, {\em closeness}, {\em hub score} and {\em authority score}. Among the worst performers are the {\em Structural Diversity Index}, and variants of {\em clustering coefficient}, the same as for keywords (table included in the supplementary material).

Precision-Recall curves for the keyphrase extraction task are shown in Figure~\ref{fig:keyphrase_PR}. Like their counterparts in keyword extraction, here also we used directed graphs and \emph{combined} gold standard annotations. We ranked vertices based on their {\em tf}, {\em tf-idf}, {\em degree}, {\em strength}, and {\em PageRank} (on weighted graph), using threshold values of 5\% to 100\% at an interval of 5. Note that here we used {\em tf} and {\em tf-idf} values of individual noun phrases rather than combining {\em tf-idf} values of n-grams as reported by Hasan and Ng~\shortcite{hasan2010conundrums}. As can be seen from Figure~\ref{fig:keyphrase_PR}, although {\em tf-idf} achieves the highest precision for the same recall when compared with other rankings on INSPEC, it is {\em degree} and {\em strength} that achieve the highest precision for the same recall when it comes to ICSI. On NUS and SemEval, performance of most centrality measures is similar to each other.


\section{Conclusion and Discussion}
\label{sec:conclusion}

In this paper, we presented a systematic study of keyword and keyphrase extraction using centrality measures on word and noun phrase collocation networks, using eleven different centrality measures and their variations. We tested our approach on four benchmark datasets that cover various domains from academic papers to meeting transcripts. We found that variants of {\em degree}, {\em strength}, {\em neighborhood size (order 1)}, and {\em PageRank} were among the top performers. In keyphrase extraction, our fast and knowledge-base-agnostic centrality-based methods outperformed the strong {\em tf-idf} baseline, and the reported results of TextRank, SingleRank, ExpandRank, and KeyCluster algorithms~\cite{hasan2010conundrums}. 

We conclude that centrality-based rankings in word and noun phrase collocation networks can be successfully used for key term extraction, without the need for a large external corpus to reliably compute the {\em idf} component of {\em tf-idf}. Furthermore, simpler centrality measures such as \emph{degree} and \emph{strength} outperform more complex and computationally more expensive measures like \emph{coreness} and \emph{betweenness}.

Interesting directions for future work consist of exploring different types of collocation networks with different window sizes, stemming words and noun phrases before adding them to our networks, taking part-of-speech tags into account, and using other centrality measures like harmonic centrality, Lin centrality~\cite{lin1976foundations}, Katz centrality~\cite{leo_katz_centrality}, and higher-order neighborhood sizes. We also hope that our study will spur further research in using centrality measures as features for supervised keyphrase extraction and rank aggregation (cf.~\cite{Liu:2007:SRA:1242572.1242638,Klementiev:2008:URA:1390156.1390216}) for key term extraction.


\bibliographystyle{acl}
\bibliography{forpaper}

\end{document}